\documentclass{article}

\usepackage{arxiv}

\usepackage[utf8]{inputenc} 
\usepackage[T1]{fontenc}    
\usepackage{hyperref}       
\usepackage{url}            
\usepackage{booktabs}       
\usepackage{amsfonts}       
\usepackage{nicefrac}       
\usepackage{microtype}      
\usepackage{xcolor}         

\usepackage{amsmath}
\usepackage{amsthm}
\usepackage{algorithm}
\usepackage{algpseudocode}
\usepackage{enumitem}
\usepackage{graphicx}
\usepackage{subcaption}
\usepackage{caption}
\usepackage{mathtools}

\newtheorem{definition}{Definition}

\newtheorem{remark}{Remark}

\DeclareMathOperator*{\argmax}{arg\,max}
\DeclareMathOperator*{\argmin}{arg\,min}

\title{Contracting with an Agent under Model Mismatch}

\author{ Elif Yılmaz \\ University of Neuchâtel \\ Neuchâtel, Switzerland \\ \texttt{elif.yilmaz@unine.ch} \And Jakub Tłuczek \\ University of Neuchâtel \\ Neuchâtel, Switzerland \\ \texttt{jakub.tluczek@unine.ch} 
\AND \hspace{0.6cm} Victor Villin \\ \hspace{0.6cm} University of Neuchâtel \\ \hspace{0.6cm} Neuchâtel, Switzerland \\ \hspace{0.6cm} \texttt{victor.villin@unine.ch}  \And Christos Dimitrakakis \\ University of Neuchâtel \\ Neuchâtel, Switzerland \\ \texttt{christos.dimitrakakis@unine.ch}   }

\begin{document}

\maketitle

\begin{abstract}
  In the classical principal-agent setting, the principal delegates some task to the agent, who receives a payment conditioned on the outcome of the task, which in turn benefits the principal. To align their objectives, the principal designs a contract that incentivizes the agent to take actions beneficial to the principal. In practice, however, the agent may act under misspecified beliefs about the underlying environment, leading to systematically suboptimal decisions. In this work, we study this problem through a principal-agent Markov Decision Process (PAMDP), in which the principal repeatedly interacts with an agent whose transition model may differ from the true environment dynamics. The principal has access to the true transition model and designs contracts that account for the agent's belief and incentives. We characterize contract design under model mismatch and develop a framework that enables the principal to induce desirable long-term benefits even though the agent has a potentially inaccurate understanding of the environment.
\end{abstract}


\section{Introduction}
The incentive alignment between two parties has long been described through the principal-agent framework, in which a principal delegates a task to an agent by offering a contract that specifies an outcome-based payment. The design of optimal contracts has been explored by economists and computer scientists \cite{bolton_contract_2005, castiglioni_multi-agent_2023, dutting_algorithmic_2024}, where it plays a fundamental role in understanding strategic interactions and incentive mechanisms in both macro- and micro-economic settings.

In the era of increasing human-AI cooperation, the contract design problem has become more relevant than ever before. When delegating tasks to AI systems, we must account for the potential misalignment of the environment dynamics, where one party cannot always assume that the other possesses an accurate model of the environment and may act according to a misspecified understanding of the environment dynamics. This form of model disagreement has been studied in a two-agent sequential stochastic game, where agents may use different models of the world while sharing the same reward structure \cite{dimitrakakis_multi-view_2017}. Existing formulations of the principal-agent Markov Decision Process (PAMDP) \cite{ben-porat_principal-agent_2023, wu_contractual_2024} assume that both parties share the same model of the world. We, however, consider a case where the agent acts according to an \emph{unknown}, potentially suboptimal \emph{misspecified} model. Under this \emph{model misalignment}, the principal's goal is to design contracts that maximize its utility while remaining incentive compatible with respect to the agent's transition model.

We study the problem of contract design under uncertainty about the agent's environment model. Our approach is based on balancing \emph{exploration}, in which the principal proposes contracts that reveal information about the agent's model, and \emph{exploitation}, in which the principal leverages the information to offer approximately optimal contracts.
Exploration is crucial because the agent's dynamics influence both the actions induced by a contract and the decision to accept it. Consequently, a principal who ignores these belief misalignment risks may design suboptimal contracts, resulting in suboptimal utility or even contract rejection.

We propose a Bayesian learning approach for contract design in PAMDPs with misspecified agent beliefs, where the principal knows the true transition dynamics and the agent uses a different model. We assume that the principal has access to a finite set of possible agent models. The principal maintains a posterior distribution over this set and updates this belief through their repeated interactions. Using the estimated agent's transition model, the principal computes least payment contracts using backward induction and linear programming. We analyze how mismatched transition dynamics in a PAMDP setting influence contract design and the resulting incentive mechanism. Our main contributions are reformulating the optimal solutions for this setting and validating our approach through empirical experiments. We further investigate how the choice of the candidate model set influences the learning process, including the accuracy of the agent's transition model estimation, the convergence of the posterior and the resulting contract design.

\subsection{Related Work}

Principal-agent problems, which formalize settings in which a principal delegates a task to an agent under moral hazard, have been extensively studied in economics and computer science as a fundamental framework \cite{bolton_contract_2005, dutting_algorithmic_2024, 
gan_generalized_2024,
grossman_hart_lp_1983,  laffont_theory_2009}. \cite{bolton_contract_2005} has provided an extensive overview of standard contract theory and incentive mechanism design. Recently, algorithmic contract design has been widely studied in the literature to investigate how contracts can be designed using computational methods \cite{alon_bayesian_2023, concha_algorithmic_2024, dutting_multi-agent_2024, dutting_algorithmic_2024,  dutting_simple_2019, guruganesh_contracts_2021}. The complexity of finding optimal contracts has been studied in \cite{bacchiocchi2024learning, dutting_complexity_2020, guruganesh_contracting_2024, zhu2022sample}. Moreover, \cite{castiglioni_multi-agent_2023, dutting_multi-agent_2022} have considered multi-agent settings where the principal designs contracts for multiple agents by extending the single agent contract design approaches. 

Optimal contract design is closely related to online learning when the principal does not have perfect information about the agent. In such settings, learning effective incentives over time through repeated interactions has become crucial in the literature \cite{dogan_estimating_2023, dogan_repeated_2023, ho_adaptive_2014, zhu_online_2023}. Furthermore, \cite{bollini_contracting_2024, guruganesh_contracting_2024, lin_generalized_2025} have focused on the settings where the agent learns over time, investigating how to design suitable contracts by incentivizing the learning agent. 

We focus on a PAMDP setting, where the principal learns how to design contracts to incentivize an agent making sequential decisions in an MDP. The principal-agent reinforcement learning framework has been introduced in \cite{ivanov_principal-agent_2024}. More recently, different approaches to designing optimal contracts have been investigated, including reward shaping \cite{ben-porat_principal-agent_2023} or history-dependent policies instead of Markovian policies \cite{bollini_contracting_2024}. 

Our setting is related to the contractual reinforcement learning framework~\cite{wu_contractual_2024}, but we consider a setting where the principal and the agent have different transition models, and the principal must infer this to design appropriate contracts. \cite{dimitrakakis_multi-view_2017} has introduced multi-view decision processes with two agents, where they disagree on transition probabilities and share the same reward function. Moreover, \cite{esponda2021equilibrium} has introduced the Berk-Nash Equilibrium to model decision making under misspecified beliefs. Recently, \cite{li2024dynamics} has analyzed the learning dynamics of a Bayesian agent with a finite set of misspecified beliefs and demonstrated that
the action frequencies converge to such an equilibrium for a learning agent with two actions. However, these works do not consider an MDP setting. In contrast, we study model mismatch in a principal-agent MDP and investigate how learning the agent's belief influences contract design.

\section{Problem Setting}
\begin{figure}[!ht]
    \centering
    \includegraphics[width=0.75\linewidth]{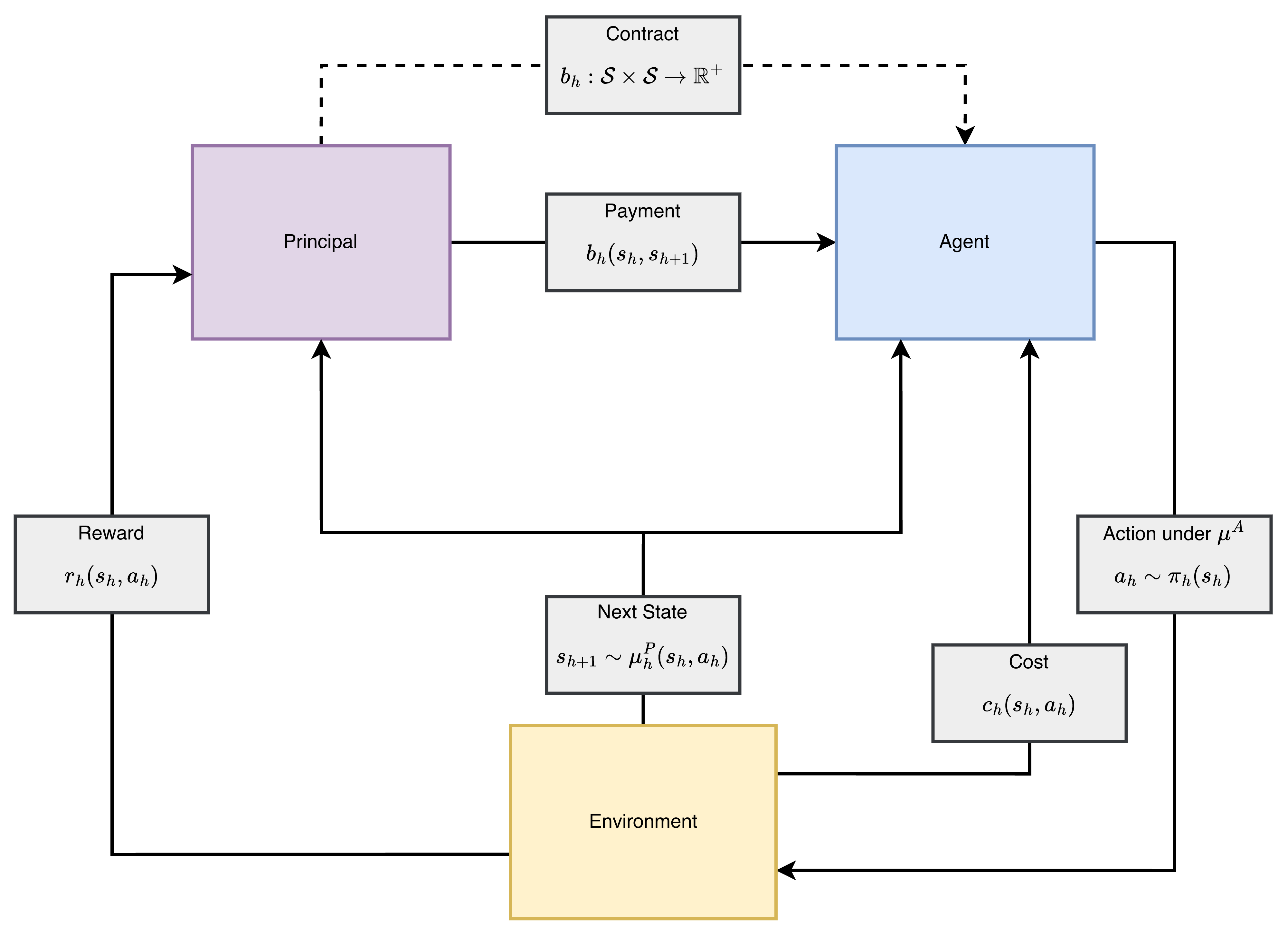}
    \caption{PAMDP setting under model mismatch}
    \label{fig:pamdp_setting}
\end{figure}

We consider a principal-agent MDP setting in which a principal interacts with an agent over a finite horizon. In each step, the principal offers a contract to the agent, and the agent selects an action based on its own transition dynamics. The resulting state transition is then realized according to the true transition dynamics of the environment. We assume that the principal has access to the true transition model while the agent may have a misspecified model. Figure \ref{fig:pamdp_setting} demonstrates how the principal and the agent interact in such an environment. 

Formally, a finite horizon Markov Decision Process (MDP) is defined by a tuple $M \coloneq (\mathcal{S}, \mathcal{A}, \{ \mu_h^P, r_h \}_{h=1}^H, \rho_0)$ where $\mathcal{S}$ is a set of possible states, $\mathcal{A}$ is an action space, and $H$ denotes the finite horizon length. In each step $h$, the transition dynamics is determined by the true transition kernel $\mu_h^P: \mathcal{S} \times \mathcal{A} \rightarrow \Delta (\mathcal{S})$, where $\mu_h^P(\cdot \mid s,a)$ specifies the probability distribution over the next state after taking action $a \in \mathcal{A}$ in state $s \in \mathcal{S}$. $r_h:\mathcal{S} \times \mathcal{A} \rightarrow \mathbb{R}$ is the environment reward function in step $h$ and the initial state is sampled from the initial state distribution $\rho_0 \in \Delta(\mathcal{S})$. 

A PAMDP is represented by a tuple $\mathcal{G} \coloneq (M, \mathcal{B},\{ \mu_h^A, c_h, \pi_h \}_{h=1}^H)$. In addition to the underlying MDP $M$, a PAMDP contains a contract space $\mathcal{B}$ such that $b_h: \mathcal{S} \times \mathcal{S} \rightarrow \mathbb{R}$ specifies the contract payment received by the agent based on the state and the next state. In each step $h$, the principal observes the current state $s_h$ and offers a contract $b_h$ to the agent. The agent then selects an action $a_h \in \mathcal{A}$ according to its policy $\pi_h: \mathcal{S} \rightarrow \Delta(\mathcal{A})$. While the agent makes decisions based on its own transition model $\mu_h^A: \mathcal{S} \times \mathcal{A} \rightarrow \Delta(\mathcal{S})$, the next state is realized according to the true transition model $\mu_h^P$. The agent suffers a cost $c_h: \mathcal{S} \times \mathcal{A} \rightarrow \mathbb{R}$ and the principal receives the reward $r_h$ defined in the underlying MDP. Thus, the principal designs contracts while accounting for the mismatch between the agent's transition model $\mu_h^A$ and the true transition model $\mu_h^P$.

\subsection{Principal and Agent Value Functions}
Given an action policy $\pi_h$ and a contract $b_h$ in a step $h$, we define value functions for the principal and the agent as they optimize different objectives and have different beliefs about the transition dynamics. The principal evaluates the expected cumulative reward under the true transition model $\mu^P$. In contrast, the agent evaluates its expected cumulative utility under its own transition model $\mu^A$, considering the contract payments received and the costs of action incurred.

We define the principal's state value function at step $h$ under any action policy $\pi$ and contract $b$ as:
\begin{align}
    & V_h^P(s \mid \pi, b) \mathbb{E}\left[ \sum_{\tau=h}^{H} r_\tau(s_\tau, a_\tau) - b_\tau(s_{\tau}, s_{\tau+1}) \mid \{ \pi_\tau \}_{\tau=h}^{H}, s_\tau = s, \mu^P  \right]  = \sum_{a \in \mathcal{A}} \pi_h (a \mid s) Q_h^P(s, a \mid \pi, b)
\end{align}
where $Q_h^P(s, a \mid \pi, b)$ is the principal's state-action value function which is written as
\begin{align}
    Q_h^P(s, a \mid \pi, b) & = \mathbb{E}\left[ \sum_{\tau=h}^{H} r_\tau(s_\tau, a_\tau) - b_\tau(s_{\tau}, s_{\tau+1})) \mid  \{ \pi_\tau \}_{\tau=h}^{H}, s_\tau = s, a_\tau = a, \mu^P  \right] \\
    & = r_h(s, a) +  \sum_{s'}  \mu_h^P(s' \mid s, a)  \left( V_{h+1}^P(s') - b_h(s, s') \right)
\end{align}
Similarly, the agent's state value function is defined by:
\begin{align}
    &V_h^A(s \mid \pi, b)  
    = \mathbb{E}\left[ \sum_{\tau=h}^{H} b_\tau(s_{\tau}, s_{\tau+1}) - c_\tau(s_\tau, a_\tau) \mid \{ \pi_\tau \}_{\tau=h}^{H}, s_\tau = s, \mu^A \right]  = \sum_{a \in \mathcal{A}} \pi_h (a \mid s) Q_h^A(s, a \mid \pi, b)
\end{align}
where $Q_h^A(s, a \mid \pi, b)$ is the agent's state-action value function which is expressed by
\begin{align}
    Q_h^A(s, a \mid \pi, b) & = \mathbb{E}\left[ \sum_{\tau=h}^{H} b_\tau(s_{\tau}, s_{\tau+1}) - c_\tau(s_\tau, a_\tau) \mid \{ \pi_\tau \}_{\tau=h}^{H}, s_\tau = s, a_\tau = a, \mu^A \right] \\
    & =  - c_h(s, a) + \sum_{s'}  \mu_h^A(s' \mid s, a) \left( V_{h+1}^A(s') + b_h(s, s') \right)
\end{align}

\subsection{Contract Constraints}
The optimal contract design problem is subject to several standard constraints containing Individual Rationality (IR), Incentive Compatibility (IC) and Limited Liability (LL). These constraints ensure that the designed contract is both acceptable to the agent and effective in inducing the desired behavior.

\begin{definition}[Individual Rationality]
    A contract $b$ satisfies the IR constraint if
    \begin{align}
        V_h^A( s \mid \pi, b) \geq 0.
    \end{align}
    Otherwise, the agent rejects the contract. It means that the agent accepts the contract only if the expected utility obtained from the induced policy is non-negative.
\end{definition}

\begin{definition}[Incentive Compatibility]
    A contract is incentive compatible if it induces the agent to take the desired action by making the intended action optimal under the agent's own model. Specifically, for a target action $a^*$ in state $s$, a contract $b$ satisfies the IC constraint if
    \begin{align}
        Q_h^A(s, a^* \mid \pi, b) \geq Q_h^A(s, a \mid \pi, b), \ \ \forall a \in \mathcal{A}. 
    \end{align}
    This constraint ensures that, given the contract and the agent's transition model $\mu^A$, the agent's best response is the desired action $a^*$.
\end{definition}

\begin{definition}[Limited liability]
     The principal cannot assign negative payments to the agent:
     \begin{align}
         b_h(s, s') \geq 0,\quad \forall s, s' \in \mathcal{S}
     \end{align}
\end{definition}

\begin{remark}
    The agent is assumed to always participate in the game. This allows us to ignore the IR constraint by adding a \emph{terminate} action $a_0$ to $\mathcal{A}$ with $r_h(s, a_0)= c_h(s, a_0)=0$ for all $s \in \mathcal{S}$ and all step $h$, which ensures nonnegative utility for the agent. Therefore, we only need to enforce the IC constraint. 
\end{remark}

Since the principal and the agent interact at each step, as illustrated in Figure \ref{fig:pamdp_setting}, selecting the terminate action at any step $h$ can be interpreted as the agent rejecting the contract for that step. In this case, the agent receives no payment and incurs no additional cost, ensuring a nonnegative utility for the agent.

\section{Optimal Solutions Using Backward Induction and Linear Program}
\label{sec:optimal_solns}
In this section, we consider a solution approach similar to \cite{wu_contractual_2024}, where the contract parameters are computed iteratively using backward induction and linear program from the final step $H$ to the initial step $1$. Unlike existing formulations, we reformulate the least payment equations under different beliefs about the transition dynamics. The goal of the optimization problem is to minimize the total contract payment of the principal subject to the agent's incentive compatibility constraint. Thus, we derive a least payment solution under the assumption of full observability of the system.

In each step $h$, the principal designs a contract that maximizes its expected utility under the true transition dynamics $\mu^P$:
\begin{align}
    b_h(s) = \argmax_{b \in \mathcal{B}} V_h^P(s \mid \pi, b)
\end{align}
Then, given a contract $b_h$, the agent selects an action according to its policy $\pi_h$ under its own transition model $\mu^A$:
\begin{align}
    \pi_h(s) = \argmax_{a \in \mathcal{A}} Q_h^A( s, a \mid \pi, b)
\end{align}
Since the principal cannot directly incentivize the agent's action, the contract is designed to induce the desired action through the agent's best response policy.

We first consider the terminal step $H$, which serves as the base case of the backward induction. We then derive the recursive formulation for the remaining steps $h < H$. Accordingly, we set $V_{H+1}^P(s) = 0$ and $V_{H+1}^A(s) = 0$. Thus, the agent's and principal's state value functions at step $H$ are given by
\begin{align}
    V_H^A(s \mid \pi, b) = \sum_{a \in \mathcal{A}} \pi_H (a \mid s) \ Q_{H}^A(s, a \mid b)
\end{align}
with 
\begin{align}
    Q_{H}^A(s, a \mid \pi, b) = - c_H(s, a) + \sum_{s' \in \mathcal{S}} \mu_H^A(s' \mid s, a) b_H(s, s') ,
\end{align}
and 
\begin{align}
    V_H^P(s \mid \pi, b) = \sum_{a \in \mathcal{A}} \pi_H (a \mid s) \ Q_{H}^P(s, a \mid b) 
\end{align}
with
\begin{align}
    Q_{H}^P(s, a \mid \pi, b) = r_H(s, a) -  \sum_{s' \in \mathcal{S}} \mu_H^P(s' \mid s, a) b_H(s, s').
\end{align}

In step $H$, for each action $a \in \mathcal{A}$, we calculate the optimal contract $b_H^*(s;a)$, which represents the least payment contract that induces the agent to select action $a$. This contract can be computed by the following linear program:
\begin{align}
    & b_H^*(s; a) = \argmin_{b \in \mathcal{B}} \sum_{s'} \mu_H^P(s' \mid s, a) b (s, s') \label{eq:optimalcontract_foractiona_atH}\\
    & \text{subject to} \notag \\
    & - c_H(s,a) + \sum_{s'} \mu_H^A(s' \mid s, a) b (s, s')  \geq - c_H(s,a') + \sum_{s'} \mu_H^A(s' \mid s, a') b (s, s'), \quad \forall a' \neq a, \notag \\
    & \hspace{0.25cm} b (s, s') \geq 0 \notag 
\end{align}
This optimization is equivalent to maximizing the principal's utility at step $H$ since the reward term $r_H(s, a)$ is independent of the contract $b$.

Using the resulting contract $b_H^*(s; a)$, the principal obtains the corresponding state-action value function $Q_{H}^P(s, a \mid b_H^*(s; a))$. Therefore, with ties broken in favor of the principal, the optimal action policy for the principal to induce is determined by
\begin{align}
    \pi_H^*(s) = \argmax_{a \in \mathcal{A}} Q_{H}^P(s, a \mid b_H^*(s; a))
\end{align}
We denote the optimal action under the policy $\pi_H^*(s)$ by $a_H^*$ since we have a deterministic action policy.
Then, the optimal contract $b_H^*(s)$ under the action $a_H^*$ is subsequently obtained by
\begin{align}
    b_H^*(s) = b_H^*(s; a^*_H).
\label{eq:optimalcontract_atH}
\end{align}
The optimal value functions of the principal and the agent in step $H$, ${V_{H}^{P^*}}$ and ${V_{H}^{A^*}}$, are then obtained under the best action policy $\pi_H^*$ and the optimal contract $b_H^*$, respectively.

At each step $h<H$, following the backward induction procedure, the principal computes the least payment contract required to induce action $a$ by solving the following:
\begin{align}
    & b_h^*(s;a) = \argmin_{b \in \mathcal{B}} \sum_{s'} \mu_h^P(s' \mid s, a) b (s, s') \label{eq:optimalcontract_foractiona_ath}\\
    & \text{subject to} \notag \\
    & - c_h(s,a) + \sum_{s'} \mu_h^A(s' \mid s, a) ({V_{h+1}^{A^*}}(s') + b (s, s')) \geq - c_h(s,a') + \sum_{s'} \mu_h^A(s' \mid s, a') ({V_{h+1}^{A^*}}(s') + b (s, s')),  \forall a' \neq a \notag  \\
    & \hspace{0.25cm} b (s, s') \geq 0 \notag 
\end{align}
where ${V_{h+1}^{A^*}}$ is under the optimal policy $\pi_{h+1}^*$ and the optimal contract $b_{h+1}^*$. 
Thus, using the principal's state-action value function $Q_{h}^P(s, a \mid b_h^*(s; a))$ under the contract $b_h^*(s; a)$, the best action policy for the principal to induce an action is:
\begin{align}
    \pi_h^*(s) = \argmax_{a \in \mathcal{A}} Q_{h}^P(s, a \mid b_h^*(s; a))
\end{align}
and the optimal contract $b_h^*(s)$ under the action $a_h^* \sim \pi_h^*(s)$ can be determined by
\begin{align}
    b_h^*(s) = b_h^*(s; a^*_h) .
\label{eq:optimalcontract_ath}
\end{align}
Similarly, in step $h$, the optimal value functions can be written using the best action policy $\pi_h^*$ and the optimal contract $b_h^*$. As a result, we note that the principal can design a contract which is likely to incorporate the agent's action costs and transition model, which enables the principal to incentivize the agent directly with the contract.

The optimal solutions are computed under the perfect information about the agent's transition model to induce a target action policy. Therefore, we note that the resulting contracts are incentive compatible, as the principal's induced policy is aligned with the best response policy of the agent.

\subsection{Motivation}
\begin{figure}[!ht]
    \centering
    \includegraphics[width=1\linewidth]{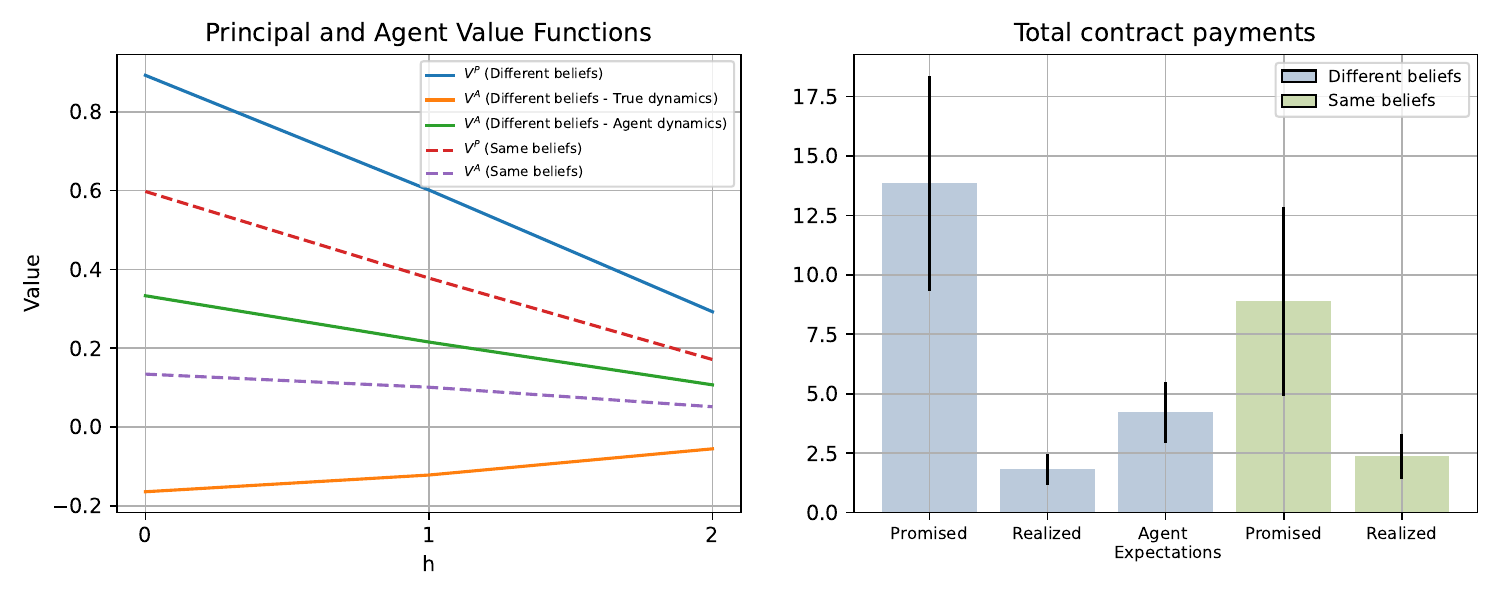}
    \caption{Comparison of optimal solutions under the same and different beliefs}
    \label{fig:diff_vs_same_beliefs}
\end{figure}

Figure \ref{fig:diff_vs_same_beliefs} compares the value functions of both the principal and the agent under settings with shared and mismatched beliefs and the corresponding total payments offered by the principal. We apply the optimal solution derived in the previous section. The results demonstrate how misspecified beliefs about the transition dynamics influence their value functions and the total contract payments by the principal. We report the average results over $20$ trials, where rewards and costs are randomly sampled from $[0, 1]$, and contract payments are bounded in $[0, 2]$. We use an MDP with $|\mathcal{S}| = 5, |\mathcal{A}| = 3, H=3$ and a uniform initial state distribution. 

We compare the total contract payments under different beliefs and the same beliefs. The promised payment is computed as the sum of all entries in the contract payment matrix. We also evaluate the expected total payment under both the agent's transition model and the true transition dynamics by accounting for the action policy. This enables us to compare the payment the agent expects to receive based on its own model and the payment the agent actually receives under the true dynamics. For their expected utility, we focus on the state value function, where we apply the backward induction steps in the previous section.

The results in Figure \ref{fig:diff_vs_same_beliefs} show that when the principal and the agent share the same belief about the transition model, the principal can induce the desired behavior with relatively lower contract payments. In contrast, under different beliefs, the principal offers higher contract payments to compensate for the discrepancy between the agent's model and the true dynamics. The figure highlights the difference between the agent's expectation and realized performance. According to its own model, the agent believes to receive a higher return; however, when evaluated under the true dynamics, the realized payments and the resulting state value of the agent are lower. Consequently, the agent overestimates its expected utility due to misspecified model, at the same time, the principal achieves higher utility under belief mismatch despite offering a higher contract payment to the agent. This is because the realized payments under the true transition dynamics are lower than those in the shared belief setting. These results demonstrate that different beliefs can significantly affect the contract design and the incentive mechanism in sequential decision-making.

\section{Learning in a PAMDP Setting with Misspecified Model}

The principal's objective is to offer contracts that incentivize the agent to act in a way that maximizes the principal's utility. We consider an episodic learning consisting of $T$ episodes, where the principal and the agent interact over $H$ steps in each episode. In each episode $t \in [T]$, at each step $h \in [H]$, the principal offers a contract and observes the agent's response, where the agent is assumed to be best responding to the offered contract according to its own  transition model. Since the state transitions are generated by the true transition dynamics $\mu^P$ (rather than $\mu^A$), the agent's model is not directly observable from the realized transitions. The principal commits to a contract payment $b$, which maximizes its expected utility, at the beginning of each episode, since the principal can anticipate the agent's best response to the committed contract. The agent computes the optimal action policy after the principal offers the contract. This implies that updates are performed at the end of each episode.

To simplify the analysis, we assume that the principal has access to a finite set of possible agent models $\mathcal{M}$. The principal knows the true transition dynamics $\mu^P$, observes $(s_h, a_h, s_{h+1})$ in each step $h$, which is stored in the trajectory $\xi_t$ for episode $t$, where $s_h$ is the current state, $a_h$ is the action taken by the agent according to the policy $\pi_h$,  which is determined by the model of the agent, and $s_{h+1}$ is the next state realized according to the true dynamics $\mu_h^P$. In addition, for any step $h$, the action cost function $c_h$ is assumed to be known by the principal. 

\begin{algorithm}
\caption{Bayesian Learning for Contract Design under Misspecified Model}
\label{alg:algo}
\begin{algorithmic}[1]

\State \textbf{Input:} $\mathcal{S}, \mathcal{A}, \mu^P, \rho_0, \mathcal{M}, H, T$

\State Initialize $P_0=\mathrm{Unif}(\mathcal{M})$.

\For{$t=1, 2, \cdots, T$}

    \State Sample $\hat{\mu}^A \sim P_t$

    \State Compute $b^t=\{b_h^t\}_{h=1}^H$ using \eqref{eq:contract_estimate} 

    \State Initialize the trajectory $\xi_t \leftarrow \emptyset$.
    
    \State Sample initial state $s_0 \sim \rho_0$
    
    \For{$h=0, 1, \cdots, H-1$}

        \State Offer $b_h^t(s_h)$ to the agent based on current state $s_h$

        \State The agent selects an action $a_{h}^t \sim \pi_h^t(s_h^t)$

        \State The agent incurs the cost $c_h(s_h^t, a_h^t)$

        \State The next state $s_{h+1}^t \sim \mu_h^P(s_h, a_h)$ is realized

        \State Observe $s_{h+1}^t$ and make the payment $b_h^t(s_h^t, s_{h+1}^t)$  

        \State Observe $a_h^t$ and receive reward $r_h(s_h^t, a_h^t)$

        \State Update $\xi_t \leftarrow \xi_t \cup \{ (s_h^t, a_h^t, s_{h+1}^t) \}$

    \EndFor

    \State Update $P_{t+1}$ using Bayes' rule based on $\xi_t$

\EndFor

\end{algorithmic}
\end{algorithm}

Therefore, the principal maintains an estimate $\hat{\mu}^A$ of the agent's transition model and updates it through repeated interaction using a Bayesian learning approach. Based on the posterior, the principal obtains an estimate $\hat{\mu}^A$ of the agent's transition model and designs contracts accordingly. In particular, the principal uses the observed trajectories in each episode to update its posterior over $\mathcal{M}$. Using the updated belief, the principal can offer approximately optimal contracts according to the true dynamics $\mu^P$, but is constrained by the estimated agent's model $\hat{\mu}^A$. Therefore, for step $h$ in episode $t$, the least payment contract is computed using estimates as follows:
\begin{align}
    & \hat{b}_h(s;a) = \argmin_{b \in \mathcal{B}} \sum_{s'} \mu_h^P(s' \mid s, a) b(s, s') \label{eq:contract_estimate}\\
    & \text{subject to} \notag \\
    & - c_h(s,a) + \sum_{s'} \hat{\mu}_h^A(s' \mid s, a) ({\hat{V}_{h+1}^A}(s') + b(s, s')) \geq - c_h(s,a') + \sum_{s'} \hat{\mu}_h^A(s' \mid s, a') ({\hat{V}_{h+1}^A}(s') + b(s, s')), \ \forall a' \neq a \notag \\
    & \hspace{0.25cm} b (s, s') \geq 0 \notag 
\end{align}
where $\hat{V}_{h}^A$ is an estimate of the agent's value under the estimated induced policy and the committed contract. Then, we follow similar steps as in Section \ref{sec:optimal_solns} to determine the best contract payment and the policy to induce for the principal.

Let $P_t$ denote the distribution over the set $\mathcal{M}$ in episode $t$. At the beginning of the interaction, a prior $P_0$ is initialized as the uniform distribution over $\mathcal{M}$. In each episode $t$, the principal updates the posterior distribution $P_t$ using observed trajectories, samples a model $\hat{\mu}^A \sim P_t$ from $\mathcal{M}$ and designs the contract which is constrained by $\hat{\mu}^A$. The learning procedure for each episode consists of the steps in Algorithm \ref{alg:algo}.

Based on the trajectory $\xi_t = \{ (s_{h}^t, a_{h}^t, s_{h+1}^t) \}_{h=1}^{H}$ in episode $t$, the principal updates its posterior belief over the possible agent models using Bayes' rule as follows:
\begin{align}
    P_{t+1}(\hat{\mu}^A) \propto P_t(\hat{\mu}^A) P(\xi_t \mid \hat{\mu}^A)
\end{align}
Since transitions to the next states are realized according to the true dynamics $\mu^P$, the observed transitions do not provide crucial information to distinguish possible agent models. Instead, the agent's model only affects the action policy. Thus, in each episode $t$, after computing the policy induced by each possible model $\hat{\mu}^A \in \mathcal{M}$, the log-likelihood term is determined by
\begin{align}
    \log \left( P \left(\xi_t \mid \hat{\mu}^A \right) \right)
    = \sum_{h=1}^{H} \log \left(
    \hat{\pi}_h \left(a_h^t \mid s_h^t, \hat{\mu}^A \right) \right)
\end{align}
where $(s_{h}^t, a_{h}^t, s_{h+1}^t) \in \xi_t$.

\section{Experiments}

In this section, we present the simulation results. First, we introduce the evaluation metrics we use throughout the experiments to demonstrate how learning the agent's misspecified model affects the contract design and how even small differences between the estimated and agent's transition models may result in significant changes in contract payments. Specifically, we consider regret to measure the principal's loss throughout their interaction, the Kullback–Leibler (KL) divergence between estimated transition model and the agent's model to evaluate estimation accuracy, and mean squared error (MSE) between learned and optimal contracts.

We evaluate the performance using expected regret over $T$ episodes, which measures the cumulative loss of the principal because of the uncertainty about the agent's transition model compared to the optimal solutions. The expected regret can be expressed as 
\begin{align}
    R_T = \sum_{t=1}^{T} \left| {V^P}^* - V^P \left( \cdot \mid \hat{\pi}^t, b^t \right) \right|
\end{align}
where ${V^P}^*$ represents the principal's optimal value under the optimal contract $b^*$ and the optimal policy $\pi^*$, which is obtained through the steps in Section \ref{sec:optimal_solns} when the agent's transition model is known, $b^t$ denotes the contract offered by the principal and $\hat{\pi}^t$ denotes the agent policy to induce for the principal by the estimated agent's transition model $\hat{\mu}^A$ in episode $t$. We use the absolute value since the principal's realized utility is not always lower than ${V^P}^*$. As shown in Figure \ref{fig:diff_vs_same_beliefs}, misspecified beliefs may enable the principal to achieve higher utility than in the same belief setting. Moreover, under model mismatch, a contract computed using the estimated agent model may result in a higher principal value than a contract computed based on any other agent's transition model.

In each episode, to measure how the principal accurately estimates the agent's model, we compute the KL divergence between the agent's true transition model $\mu^A$ and the estimated one $\hat{\mu}^A$. For every state action pair $(s,a)$, we have the next state distribution and compute the average KL divergence over all pairs as
\begin{align}
    & D_{\operatorname{KL}}\left( \mu^A \mid \mid \hat{\mu}^A  \right)  = \frac{1}{H |\mathcal{A}| |\mathcal{S}| } \sum_{h \in [H]} \sum_{s, s' \in \mathcal{S}} \sum_{a \in \mathcal{A}} \mu_h^A(s' \mid s, a) \log \left( \frac{\mu_h^A(s' \mid s, a)}{\hat{\mu}_h^A(s' \mid s, a)} \right)
\end{align}
where $|\mathcal{S}|$ and $|\mathcal{A}|$ are the number of states and actions, respectively. 

Furthermore, since the principal updates its posterior estimate over the set of possible agent models, we also evaluate the KL divergence between the agent's and the estimated models using the posterior distributions over the set. Thus, the KL divergence with the posterior is defined as follows:
\begin{align}
    & D_{\operatorname{KL}}^{\text{posterior}}\left( \mu^A \mid \mid \mathcal{M}  \right)  = \frac{1}{H |\mathcal{A}| |\mathcal{S}| } \sum_{h \in [H]} \sum_{s, s' \in \mathcal{S}} \sum_{a \in \mathcal{A}} \mu_h^A(s' \mid s, a) \log \left( \frac{\mu_h^A(s' \mid s, a)}{ \sum_{\hat{\mu}^A \in \mathcal{M}} P_t(\hat{\mu}^A)\hat{\mu}_h^A(s' \mid s, a)} \right)
\end{align}

To evaluate how much the learned contracts differ from the optimal one, we compute the MSE between the learned contract $b$ and the optimal payments $b^*$, which is defined as
\begin{align}
    \operatorname{MSE} \left(b^*, b \right) = \frac{1}{H |\mathcal{S}|^2} \sum_{h \in [H]} \sum_{s, s' \in \mathcal{S}} \left( b_h^*(s, s') - b_h(s, s') \right)^2 
\end{align}

To construct a finite set of candidate agent transition models, we generate transition probabilities by sampling from a Dirichlet distribution centered around a combination of the agent's model and the true transition dynamics. For any $s \in \mathcal{S}$, $a \in \mathcal{A}$, and step $h$, we sample
\begin{align}
    \mu_h( \cdot \mid s, a) \sim \mathrm{Dirichlet} \left(
    \alpha\left( \varepsilon \mu_h^A(\cdot \mid s,a) + (1 -\varepsilon ) \mu_h^P( \cdot \mid s, a) \right) \right)
\end{align}
where $\alpha >0$ is the concentration parameter of the Dirichlet distribution, controlling how closely the sampled transition probabilities are concentrated around the given distribution and $\varepsilon \in [0,1]$ determines the influence of the agent's model and true dynamics. For example, if $\varepsilon = 1$, then the sampled distribution is concentrated on the agent's transition model.

We consider a PAMDP setting with $|\mathcal{S}| = 5, |\mathcal{A}| = 3, \rho_0 = \mathrm{Unif}(\mathcal{S})$ and $T=200$. The agent and true transition models are randomly generated and remain fixed throughout the experiments. The rewards and costs are also randomly generated from $[0,1]$, and the contract payments are bounded within $[0,2]$. We note that all results are averaged over $20$ trials.

\begin{figure}[!ht]
    \centering
    \includegraphics[width=\textwidth]{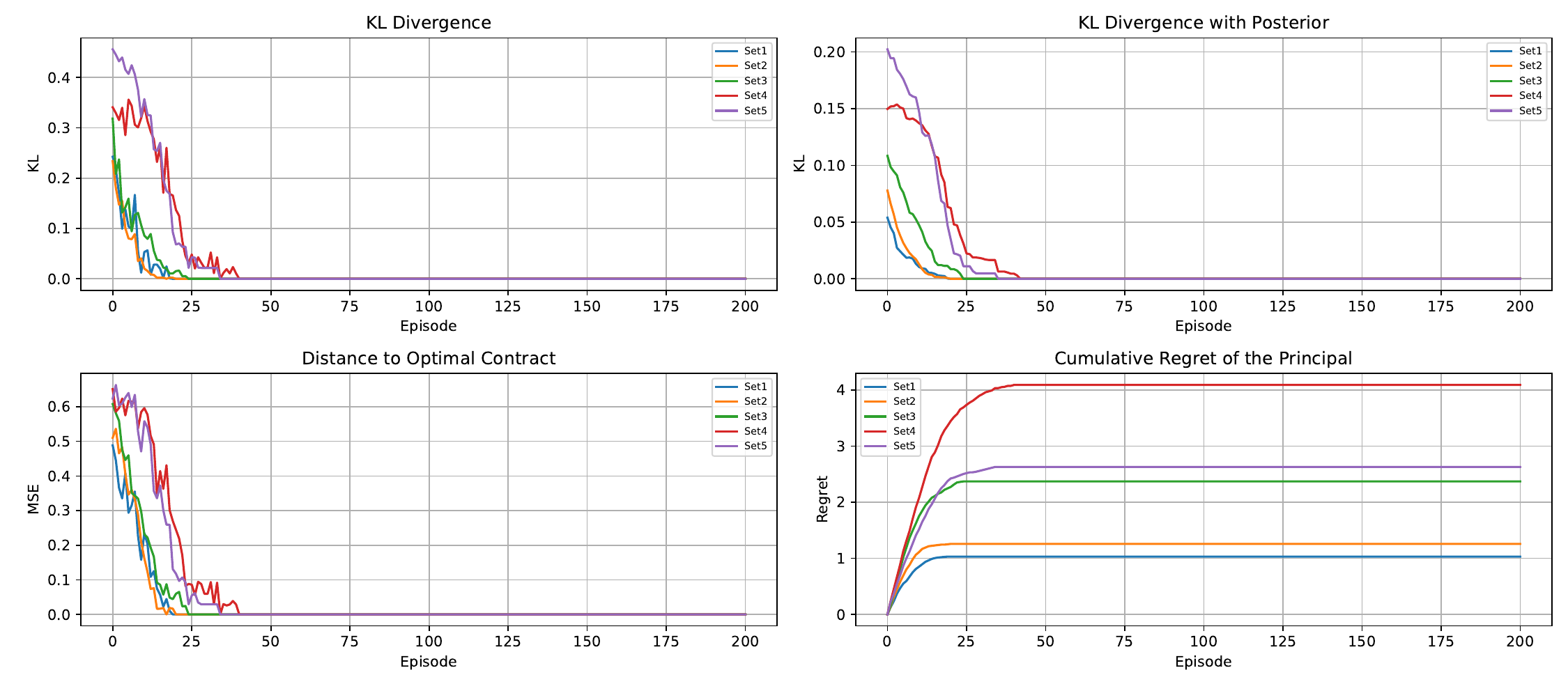}
    \caption{Comparison of different model sets containing the agent's transition model $\mu^A$ for $H=3$}
    \label{fig:exp1}
\end{figure}

\begin{figure}[!ht]
    \centering
    \includegraphics[width=\textwidth]{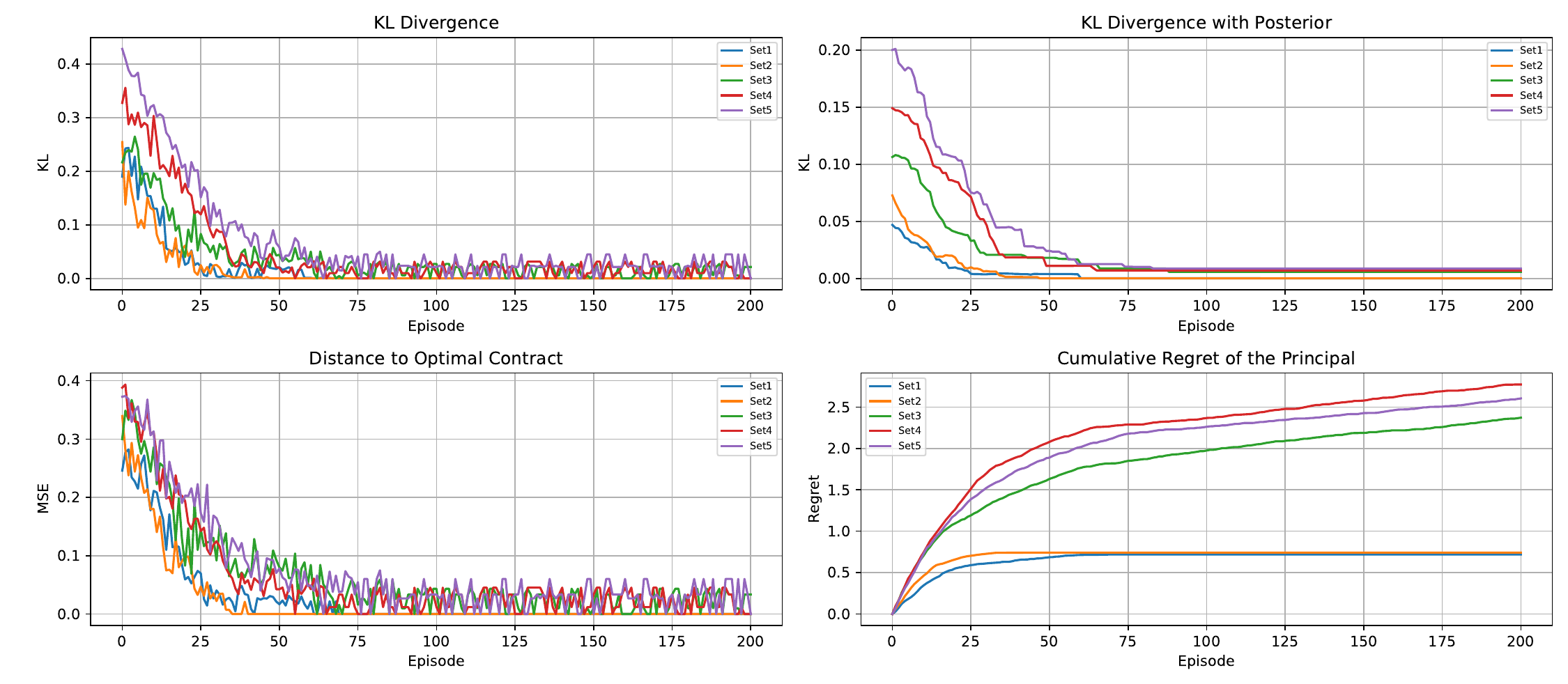}
    \caption{Comparison of different model sets containing the agent's transition model $\mu^A$ for $H=1$}
    \label{fig:exp2}
\end{figure}

Figures \ref{fig:exp1} and \ref{fig:exp2} compare the effect of the time horizon $H$ on learning the agent's transition model when all sets contain $\mu^A$ and $\mu^P$. We construct five different model sets by fixing $\alpha=10^2$ and varying the coefficient $\varepsilon \in \{1, 0.75, 0.5, 0.25, 0\}$, respectively, from Set 1 to 5. Each set contains 300 transition models, 150 randomly generated models, $\mu^A$, and $\mu^P$, while the remaining models are sampled from the Dirichlet distribution described above. We note that the average KL divergence between the true dynamics and the agent's model, $D_{KL} \left( \mu^P \mid \mid \mu^A \right)$, is 0.41 for Figure  \ref{fig:exp1} and 0.40 for Figure \ref{fig:exp2}.

In Figure \ref{fig:exp1}, we set $H=3$, whereas in Figure \ref{fig:exp2}, $H=1$. With a higher number of interactions in each episode, the principal is able to distinguish the agent's transition model more easily, and the KL divergence with the posterior consistently converges to $\mu^A$ across all sets, as shown in Figure \ref{fig:exp1}. In contrast, in Figure \ref{fig:exp2}, the principal observes only one transition per episode, which may lead to insufficient exploration of some states and actions in some trials. Since some of the state action pairs are rarely or never visited by the agent, some models may induce similar likelihoods over the observed trajectories. Consequently, when they interact for fewer steps in each episode, it may result in convergence to a suboptimal posterior by assigning a positive probability to some suboptimal models. We also observe that even small differences in transition probabilities may result in significant loss for the principal, which means that models with similar KL values may lead to different performance, while models with larger KL divergence can still achieve comparable results.

\begin{figure}[!ht]
    \centering
    \includegraphics[width=\textwidth]{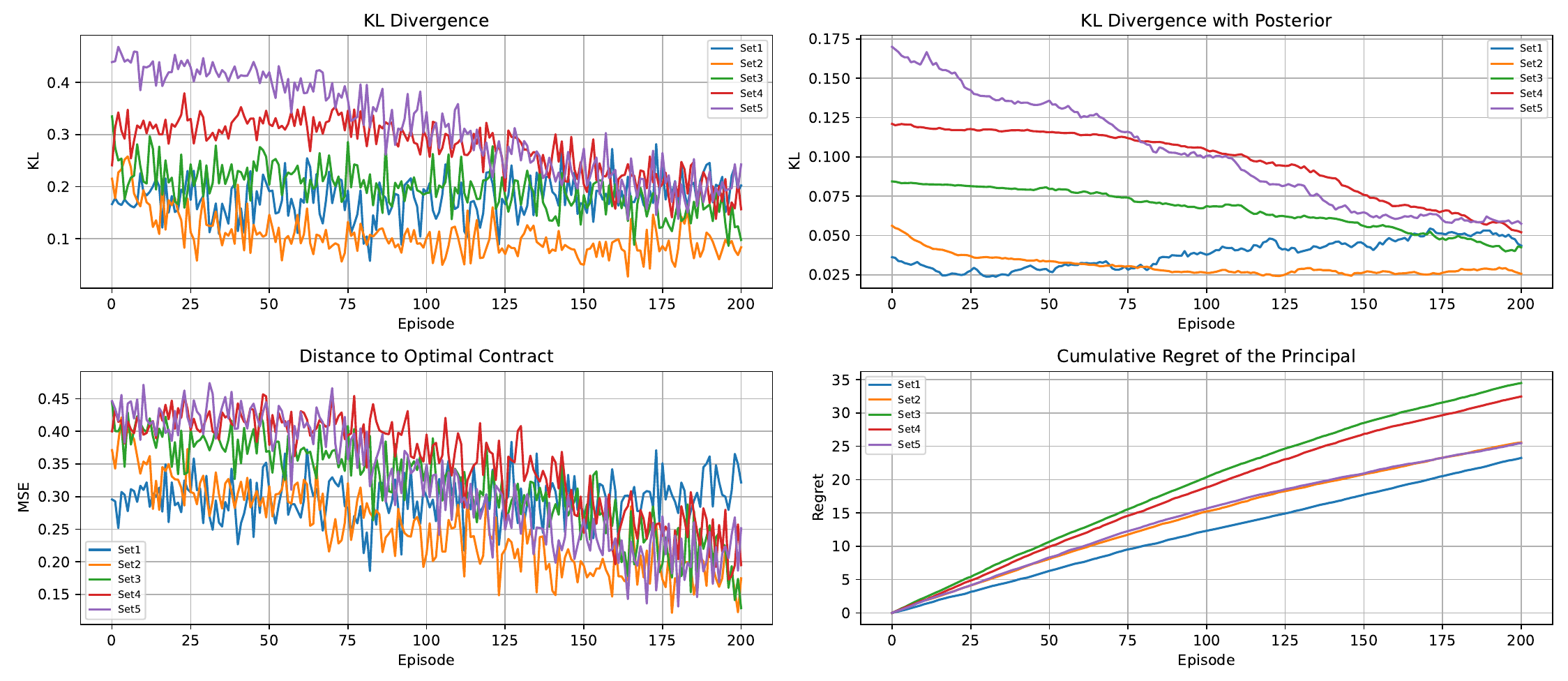}
    \caption{Comparison of different model sets that do not contain the agent's transition model $\mu^A$ for $H=3$}
    \label{fig:exp3}
\end{figure}

Furthermore, we present additional experiments for model sets which do not contain the agent's true transition dynamics $\mu^A$, as shown in Figure \ref{fig:exp3}, to investigate whether the posterior distribution converges to the model that is closest to the agent's model. We consider sets of models constructed in the same style as those in Figure \ref{fig:exp1}, except that $\mu^A$ is excluded from the sets. We note that $D_{KL} \left( \mu^P \mid \mid \mu^A \right)$ is 0.36 and the closest model to $\mu^A$ has KL divergence of 0.002 for all sets.

Since the agent's model is not in the sets, exact convergence to the agent's model, as observed in Figures \ref{fig:exp1} and \ref{fig:exp2}, is no longer possible. Thus, we update the posterior using a Boltzmann softmax policy with a temperature parameter of 1 to obtain a meaningful belief distribution over the model set while keeping the rest of the learning procedure unchanged. The results demonstrate that Set 1, which contains models closer to $\mu^A$, achieves lower regret, but also results in larger fluctuations in the contract distance and the KL divergence, since this set contains more candidate models similar to $\mu^A$, making the posterior distribution less concentrated. We observe a decreasing trend in the KL divergence and contract distance as the closest model consistently maintains a positive posterior probability. On the other hand, as shown in the KL divergence plot, multiple models maintain positive posterior probabilities throughout the learning process, preventing convergence in KL divergence and contract distance.

Consequently, when the sets contain the agent's transition model, we observe convergence in the plots as long as the principal and agent interact sufficiently. In contrast, when the agent's model is excluded from the sets, such convergence is no longer guaranteed. Our experiments show that while KL divergence and contract distance provide useful insight into the learning process, they do not fully capture the effectiveness of the learning approach. In particular, regret provides a more informative measure of performance. We also observe across all three experiments that Sets 1 and 2, which contain models closer to the agent's transition model, consistently yield lower regret for the principal. We conclude that accurately learning the agent's dynamics is essential to design effective contracts and provide incentives under belief mismatch.

\section{Discussion}
In this work, we study a PAMDP setting with model misalignment where the principal and the agent have different beliefs on transition dynamics. We investigate how this mismatch affects the contract design. In particular, we consider the setting where the principal has access to a set of possible agent transition models and designs contracts by learning the agent's model through repeated interactions. Our experimental results show that accurately estimating the agent's transition model is essential for effective contract design. As the principal's estimate converges to the agent's model, the learned contracts approach the optimal contracts computed under perfect information, and thus the principal incurs lower regret. Moreover, the experiments demonstrate that even small differences in learning the agent's model may result in noticeable changes in contract payments and the principal's loss. In future work, our aim is to extend our analysis to continuous model spaces and analyze how contracts can be used to estimate the misaligned belief of the agent. Moreover, an important extension is to establish theoretical guarantees for this learning framework. Finally, many PAMDP settings consider that the principal cannot directly observe the agent's actions. Extending our approach to such settings would be an interesting direction, as it has additional challenges in learning the agent's model and contract design.

\bibliographystyle{plain}
\bibliography{references}

\begin{thebibliography}{10}

\bibitem{alon_bayesian_2023}
Tal Alon, Paul Dütting, Yingkai Li, and Inbal Talgam-Cohen.
\newblock Bayesian {Analysis} of {Linear} {Contracts}, July 2023.
\newblock arXiv:2211.06850 [cs, econ].

\bibitem{bacchiocchi2024learning}
Francesco Bacchiocchi, Matteo Castiglioni, Alberto Marchesi, and Nicola Gatti.
\newblock Learning optimal contracts: How to exploit small action spaces.
\newblock In {\em International Conference on Learning Representations}, volume 2024, pages 11944--11970, 2024.

\bibitem{ben-porat_principal-agent_2023}
Omer Ben-Porat, Yishay Mansour, Michal Moshkovitz, and Boaz Taitler.
\newblock Principal-{Agent} {Reward} {Shaping} in {MDPs}, December 2023.
\newblock arXiv:2401.00298 [cs].

\bibitem{bollini_contracting_2024}
Matteo Bollini, Francesco Bacchiocchi, Matteo Castiglioni, Alberto Marchesi, and Nicola Gatti.
\newblock Contracting {With} a {Reinforcement} {Learning} {Agent} by {Playing} {Trick} or {Treat}, October 2024.
\newblock arXiv:2410.13520 [cs].

\bibitem{bolton_contract_2005}
Patrick Bolton and Mathias Dewatripont.
\newblock {\em Contract theory}.
\newblock MIT Press, Cambridge, MA Londin, England, 2005.

\bibitem{castiglioni_multi-agent_2023}
Matteo Castiglioni, Alberto Marchesi, and Nicola Gatti.
\newblock Multi-{Agent} {Contract} {Design}: {How} to {Commission} {Multiple} {Agents} with {Individual} {Outcomes}.
\newblock In {\em Proceedings of the 24th {ACM} {Conference} on {Economics} and {Computation}}, pages 412--448, London United Kingdom, July 2023. ACM.

\bibitem{concha_algorithmic_2024}
David~Molina Concha, Kyeonghyeon Park, Hyun-Rok Lee, Taesik Lee, and Chi-Guhn Lee.
\newblock Algorithmic {Contract} {Design} with {Reinforcement} {Learning} {Agents}, August 2024.
\newblock arXiv:2408.09686 [cs].

\bibitem{dimitrakakis_multi-view_2017}
Christos Dimitrakakis, David~C. Parkes, Goran Radanovic, and Paul Tylkin.
\newblock Multi-view decision processes: the helper-{AI} problem.
\newblock In {\em Proceedings of the 31st {International} {Conference} on {Neural} {Information} {Processing} {Systems}}, {NIPS}'17, pages 5449--5458, Red Hook, NY, USA, 2017. Curran Associates Inc.

\bibitem{dogan_estimating_2023}
Ilgin Dogan, Zuo-Jun~Max Shen, and Anil Aswani.
\newblock Estimating and {Incentivizing} {Imperfect}-{Knowledge} {Agents} with {Hidden} {Rewards}, August 2023.
\newblock arXiv:2308.06717 [cs, stat].

\bibitem{dogan_repeated_2023}
Ilgin Dogan, Zuo-Jun~Max Shen, and Anil Aswani.
\newblock Repeated {Principal}-{Agent} {Games} with {Unobserved} {Agent} {Rewards} and {Perfect}-{Knowledge} {Agents}, May 2023.
\newblock arXiv:2304.07407 [cs, stat].

\bibitem{dutting_multi-agent_2022}
Paul Dütting, Tomer Ezra, Michal Feldman, and Thomas Kesselheim.
\newblock Multi-{Agent} {Contracts}, November 2022.
\newblock arXiv:2211.05434 [cs].

\bibitem{dutting_multi-agent_2024}
Paul Dütting, Tomer Ezra, Michal Feldman, and Thomas Kesselheim.
\newblock Multi-{Agent} {Combinatorial} {Contracts}, May 2024.
\newblock arXiv:2405.08260 [cs].

\bibitem{dutting_algorithmic_2024}
Paul Dütting, Michal Feldman, and Inbal Talgam-Cohen.
\newblock Algorithmic {Contract} {Theory}: {A} {Survey}, December 2024.
\newblock arXiv:2412.16384 [cs].

\bibitem{dutting_simple_2019}
Paul Dütting, Tim Roughgarden, and Inbal Talgam-Cohen.
\newblock Simple versus {Optimal} {Contracts}.
\newblock In {\em Proceedings of the 2019 {ACM} {Conference} on {Economics} and {Computation}}, pages 369--387, Phoenix AZ USA, June 2019. ACM.

\bibitem{dutting_complexity_2020}
Paul Dütting, Tim Roughgarden, and Inbal Talgam-Cohen.
\newblock The {Complexity} of {Contracts}, February 2020.
\newblock arXiv:2002.12034 [cs].

\bibitem{esponda2021equilibrium}
Ignacio Esponda and Demian Pouzo.
\newblock Equilibrium in misspecified markov decision processes.
\newblock {\em Theoretical Economics}, 16(2):717--757, 2021.

\bibitem{gan_generalized_2024}
Jiarui Gan, Minbiao Han, Jibang Wu, and Haifeng Xu.
\newblock Generalized {Principal}-{Agency}: {Contracts}, {Information}, {Games} and {Beyond}, February 2024.
\newblock arXiv:2209.01146 [cs].

\bibitem{grossman_hart_lp_1983}
Sanford Grossman and Oliver Hart.
\newblock An analysis of principal-agent problem.
\newblock {\em Econometrica}, 51:7--45, 01 1983.

\bibitem{guruganesh_contracting_2024}
Guru Guruganesh, Yoav Kolumbus, Jon Schneider, Inbal Talgam-Cohen, Emmanouil-Vasileios Vlatakis-Gkaragkounis, Joshua~R. Wang, and S.~Matthew Weinberg.
\newblock Contracting with a {Learning} {Agent}, January 2024.
\newblock arXiv:2401.16198 [cs, econ].

\bibitem{guruganesh_contracts_2021}
Guru Guruganesh, Jon Schneider, and Joshua~R. Wang.
\newblock Contracts under {Moral} {Hazard} and {Adverse} {Selection}.
\newblock In {\em Proceedings of the 22nd {ACM} {Conference} on {Economics} and {Computation}}, pages 563--582, Budapest Hungary, July 2021. ACM.

\bibitem{ho_adaptive_2014}
Chien-Ju Ho, Aleksandrs Slivkins, and Jennifer~Wortman Vaughan.
\newblock Adaptive contract design for crowdsourcing markets: bandit algorithms for repeated principal-agent problems.
\newblock In {\em Proceedings of the fifteenth {ACM} conference on {Economics} and computation}, pages 359--376, Palo Alto California USA, June 2014. ACM.

\bibitem{ivanov_principal-agent_2024}
Dima Ivanov, Paul Dütting, Inbal Talgam-Cohen, Tonghan Wang, and David~C. Parkes.
\newblock Principal-{Agent} {Reinforcement} {Learning}, July 2024.
\newblock arXiv:2407.18074 [cs].

\bibitem{laffont_theory_2009}
Jean-Jacques Laffont and David Martimort.
\newblock {\em The {Theory} of {Incentives}: {The} {Principal}-{Agent} {Model}}.
\newblock Princeton University Press, December 2009.

\bibitem{li2024dynamics}
Yingkai Li and Argyris Oikonomou.
\newblock Dynamics and contracts for an agent with misspecified beliefs.
\newblock {\em arXiv preprint arXiv:2405.20423}, 2024.

\bibitem{lin_generalized_2025}
Tao Lin and Yiling Chen.
\newblock Generalized {Principal}-{Agent} {Problem} with a {Learning} {Agent}, October 2025.
\newblock arXiv:2402.09721 [cs].

\bibitem{wu_contractual_2024}
Jibang Wu, Siyu Chen, Mengdi Wang, Huazheng Wang, and Haifeng Xu.
\newblock Contractual {Reinforcement} {Learning}: {Pulling} {Arms} with {Invisible} {Hands}, July 2024.
\newblock arXiv:2407.01458 [cs].

\bibitem{zhu2022sample}
Banghua Zhu, Stephen Bates, Zhuoran Yang, Yixin Wang, Jiantao Jiao, and Michael~I Jordan.
\newblock The sample complexity of online contract design.
\newblock {\em arXiv preprint arXiv:2211.05732}, 2022.

\bibitem{zhu_online_2023}
Banghua Zhu, Sai~Praneeth Karimireddy, Jiantao Jiao, and Michael~I. Jordan.
\newblock Online {Learning} in a {Creator} {Economy}, May 2023.
\newblock arXiv:2305.11381 [cs].

\end{thebibliography}

\end{document}